\documentclass[a4paper]{article}
\usepackage{iwslt18,amssymb,amsmath,epsfig}
\usepackage{hyperref}
\def\reg{{\rm\ooalign{\hfil
 \raise.07ex\hbox{\scriptsize R}\hfil\crcr\mathhexbox20D}}}

\title{Transformer-based Cascaded Multimodal Speech Translation}

\usepackage{xspace}
\usepackage{parskip}
\usepackage[colorinlistoftodos]{todonotes}
\usepackage{booktabs}

\newcommand{\smallsection}[1]{\textbf{#1}\par}
\newcommand{\tensortwotensor}{\texttt{tensor2tensor}}

\newcommand{\bleu}{BLEU\xspace}

\newcommand{\textonly}{\texttt{Baseline}\xspace}
\newcommand{\videosum}{\texttt{AvgPool}\xspace}
\newcommand{\clo}{\texttt{Conv}\xspace}

\newcommand{\thace}{\texttt{Emb}\xspace}

\newcommand{\enc}{\texttt{Enc}\xspace}
\newcommand{\dec}{\texttt{Dec}\xspace}

\newcommand{\avc}{\texttt{Cond}\xspace}
\newcommand{\avf}{\texttt{Attn}\xspace}

\newcommand{\trans}{\texttt{Trans}\xspace}
\newcommand{\delib}{\texttt{Delib}\xspace}

\newcommand{\additive}{\texttt{A}\xspace}
\newcommand{\cascade}{\texttt{C}\xspace}

\newcommand{\vanilla}{\texttt{Van}\xspace}

\newcommand{\firstpass}{\texttt{1P}\xspace}
\newcommand{\secondpass}{\texttt{2P}\xspace}

\makeatletter
\def\name#1{\gdef\@name{#1\\}}
\makeatother
\name{\em {Zixiu Wu$^{1}$, Ozan Caglayan$^{1}$, Julia Ive$^{2}$, Josiah Wang$^{1}$, Lucia Specia$^{1}$}}
\address{$^{1}$Department of Computing, Imperial College London, UK \\
         $^{2}$Department of Computer Science, University of Sheffield, UK\\
{\small \tt \{zixiu.wu18, o.caglayan, josiah.wang, l.specia\}@imperial.ac.uk}\\
{\small \tt j.ive@sheffield.ac.uk}}

\begin{document}
\maketitle

\begin{abstract}
This paper describes the cascaded multimodal speech translation systems
developed by Imperial College London for the IWSLT 2019 evaluation campaign.
The architecture consists of an automatic speech recognition (ASR) system followed by a Transformer-based multimodal machine translation (MMT) system. While the ASR component is identical across the experiments, the MMT model varies in terms of the
way of integrating the visual context (simple conditioning vs. attention), the type of visual features exploited (pooled, convolutional, action categories) and the underlying architecture. For the latter, we explore both the canonical transformer \cite{vaswani2017attention} and its deliberation version \cite{hassan2018achieving} with additive and cascade variants which differ in how they integrate the textual attention.
Upon conducting extensive experiments, we found that (i) the explored visual integration schemes often harm the translation performance for the transformer and additive deliberation, but considerably improve the cascade deliberation; (ii) the transformer and cascade deliberation integrate the visual modality better than the additive deliberation, as shown by the incongruence analysis.





\end{abstract}

\section{Introduction}
\label{introduction}
The recently introduced How2 dataset \cite{how2} has stimulated research around multimodal language understanding through the availability of \textit{300h} instructional videos, English subtitles and their Portuguese translations. For example, \cite{wu2019predicting} successfully demonstrates that semantically rich action-based visual features are helpful in the context of machine translation (MT), especially in the presence of input noise that manifests itself as missing source words. Therefore, we hypothesize that a speech-to-text translation (STT) system may also benefit from the visual context, especially in the traditional \textit{cascaded} framework \cite{casacuberta_st,waibel2008spoken}
where noisy automatic transcripts are obtained from an automatic speech recognition system (ASR) and further translated into the target language using a machine translation (MT) component. The dataset enables the design of such multimodal STT systems, since we have access to a bilingual corpora as well as the corresponding audio-visual stream. Hence, in this paper, we propose a cascaded multimodal STT with two components: (i) an English ASR system trained on the How2 dataset and (ii) a transformer-based \cite{vaswani2017attention} visually grounded MMT system.

MMT is a relatively new research topic which is interested in leveraging auxiliary modalities such as audio or vision in order to improve translation performance \cite{specia2016shared}. MMT has proved effective in scenarios such as for disambiguation \cite{ive2019distil} or when the source sentences are corrupted \cite{naacl19-probing}. So far, MMT has mostly focused on integrating visual features into neural MT (NMT) systems using visual attention through convolutional feature maps \cite{caglayan-EtAl:2016:WMT,libovicky2017attention} or visual conditioning of encoder/decoder blocks through fully-connected features \cite{caglayan-EtAl:2017:WMT,calixto2017incorporating,madhyastha2017sheffield,gronroos2018memad}.

Inspired by previous research in MMT, we explore several multimodal integration schemes using action-level video features. Specifically, we experiment with visually conditioning the encoder output and adding visual attention to the decoder. We further extend the proposed schemes to the deliberation variant \cite{hassan2018achieving} of the canonical transformer in two ways: additive and cascade multimodal deliberation, which are distinct in their textual attention regimes. Overall, the results show that multimodality in general leads to performance degradation for the canonical transformer and the additive deliberation variant, but can result in substantial improvements for the cascade deliberation. Our incongruence analysis~\cite{elliott2018adversarial} reveals that the transformer and cascade deliberation are more sensitive to and therefore more reliant on visual features for translation, whereas the additive deliberation is much less impacted. We also observe that incongruence sensitivity and translation performance are not necessarily correlated.

\section{Methods}
In this section, we briefly describe the proposed multimodal speech translation system and its components.

\subsection{Automatic Speech Recognition}
The baseline ASR system that we use to obtain English transcripts is an attentive sequence-to-sequence architecture with a stacked \textbf{encoder} of 6 bidirectional LSTM layers \cite{hochreiter1997long}. Each
LSTM layer is followed by a \textit{tanh} projection layer.
The middle two LSTM layers apply a temporal subsampling \cite{las} by skipping every other input, reducing the length of the sequence $\mathrm{X}$ from $T$ to $T/4$. All LSTM and projection layers have 320 hidden units.
The forward-pass of the encoder produces the source encodings on top of which attention will be applied within the decoder. The hidden and cell states of all LSTM layers are initialized with $0$.
The \textbf{decoder} is a 2-layer stacked GRU \cite{Chung2014}, where the first GRU receives the previous hidden state of the second GRU in a transitional way. GRU layers, attention layer and embeddings have 320 hidden units. We share the input and output embeddings to reduce the number of parameters \cite{press2016using}. At timestep $t\mathrm{=}0$, the hidden state of the first GRU is initialized with the average-pooled source encoding states.

\subsection{Deliberation-based NMT}
A human translator typically produces a translation draft first, and then refines it towards the final translation. The idea behind the deliberation networks \cite{xia2017deliberation} simulates this process by extending the conventional attentive encoder-decoder architecture \cite{Bahdanau2014} with a second pass \textit{refinement} decoder.
Specifically, the encoder first encodes a source sentence of length $N$ into a sequence of hidden states $\mathcal{H} = \{h_1, h_2,\dots,h_{N}\}$ on top of which the first pass decoder \firstpass applies the attention. The pre-softmax hidden states $\{\hat{s}_1,\hat{s}_2,\dots,\hat{s}_{M}\}$ produced by the decoder leads to a first pass translation $\{\hat{y}_1,\hat{y}_2,\dots, \hat{y}_{M}\}$. The second pass decoder \secondpass intervenes at this point and generates a second translation by attending separately to both $\mathcal{H}$ and the concatenated state vectors $\{[\hat{s}_1;\hat{y}_1], [\hat{s}_2; \hat{y}_2],\dots,[\hat{s}_{M}; \hat{y}_{M}]\}$.
Two context vectors are produced as a result, and they are joint inputs with $s_{t-1}$ (previous hidden state of \secondpass) and $y_{t-1}$ (previous output of \secondpass) to \secondpass to yield $s_t$ and then $y_t$.

A transformer-based deliberation architecture is proposed by \cite{hassan2018achieving}. It follows the same two-pass refinement process, with every second-pass decoder block attending to both the encoder output $\mathcal{H}$ and the first-pass pre-softmax hidden states $\mathcal{\hat{S}}$. However, it differs from \cite{xia2017deliberation} in that the actual first-pass translation $\hat{Y}$ is not used for the second-pass attention.


\subsection{Multimodality}
\label{multimodality}

\subsubsection{Visual Features}
\label{visual features}

We experiment with three types of video features, namely average-pooled vector representations (\videosum), convolutional layer outputs (\clo), and Ten-Hot action category embeddings (\thace). The \videosum features are provided by the How2 dataset using the following approach: a video is segmented into smaller parts of 16 frames each, and the segments are fed to a 3D ResNeXt-101 CNN \cite{xie2017aggregated}, trained to recognise 400 action classes \cite{hara2018can}. The $2048$-D fully-connected features are then averaged across the segments to obtain a single \videosum feature vector for the overall video.

In order to obtain the \clo features, 16 equi-distant frames are sampled from a video, and they are then used as input to an inflated 3D ResNet-50 CNN \cite{monfortmoments} fine-tuned on the \textit{Moments in Time} action video dataset. The CNN hence takes in a video and classifies it into one of 339 categories. The \clo features, taken at the CONV$_4$ layer of the network, has a $7 \times 7 \times 2048$ dimensionality.

Higher-level semantic information can be more helpful than convolutional features. We apply the same CNN to a video as we do for \clo features, but this time the focus is on the softmax layer output: we process the embedding matrix to keep the 10 most probable category embeddings intact while zeroing out the remaining ones. We call this representation ten-hot action category embeddings (\thace).



\subsubsection{Integration Approaches}
\label{multimodal encoding and decoding}


\smallsection{Encoder with Additive Visual Conditioning (\enc-\avc)}
In this approach, inspired by \cite{ive2019distil}, we add a projection of the visual features to each output of the vanilla transformer encoder (\enc-\vanilla). This projection is strictly linear from the $2048$-D \videosum features to the $1024$-D space in which the self attention hidden states reside, and the projection matrix is learned jointly with the translation model.

\smallsection{Decoder with Visual Attention (\dec-\avf)}
In order to accommodate attention to visual features at the decoder side and inspired by \cite{helcl2018cuni}, we insert one layer of visual cross attention at a decoder block immediately before the fully-connected layer. We name the transformer decoder with such an extra layer as \trans-\dec-\avf, where this layer is immediately after the textual attention to the encoder output. Specifically,
we experiment with attention to \clo, \thace and \videosum features separately. The visual attention is distributed across the 49 video regions in \clo, the 339 action category word embeddings in \thace, or the 32 rows in \videosum where we reshape the 2048-D \videosum vector into a $32 \times 64$ matrix.

\subsubsection{Multimodal Transformers}
\label{multimodal transformers}
\begin{figure*}[ht]
\centering
\includegraphics[width=.6\textwidth]{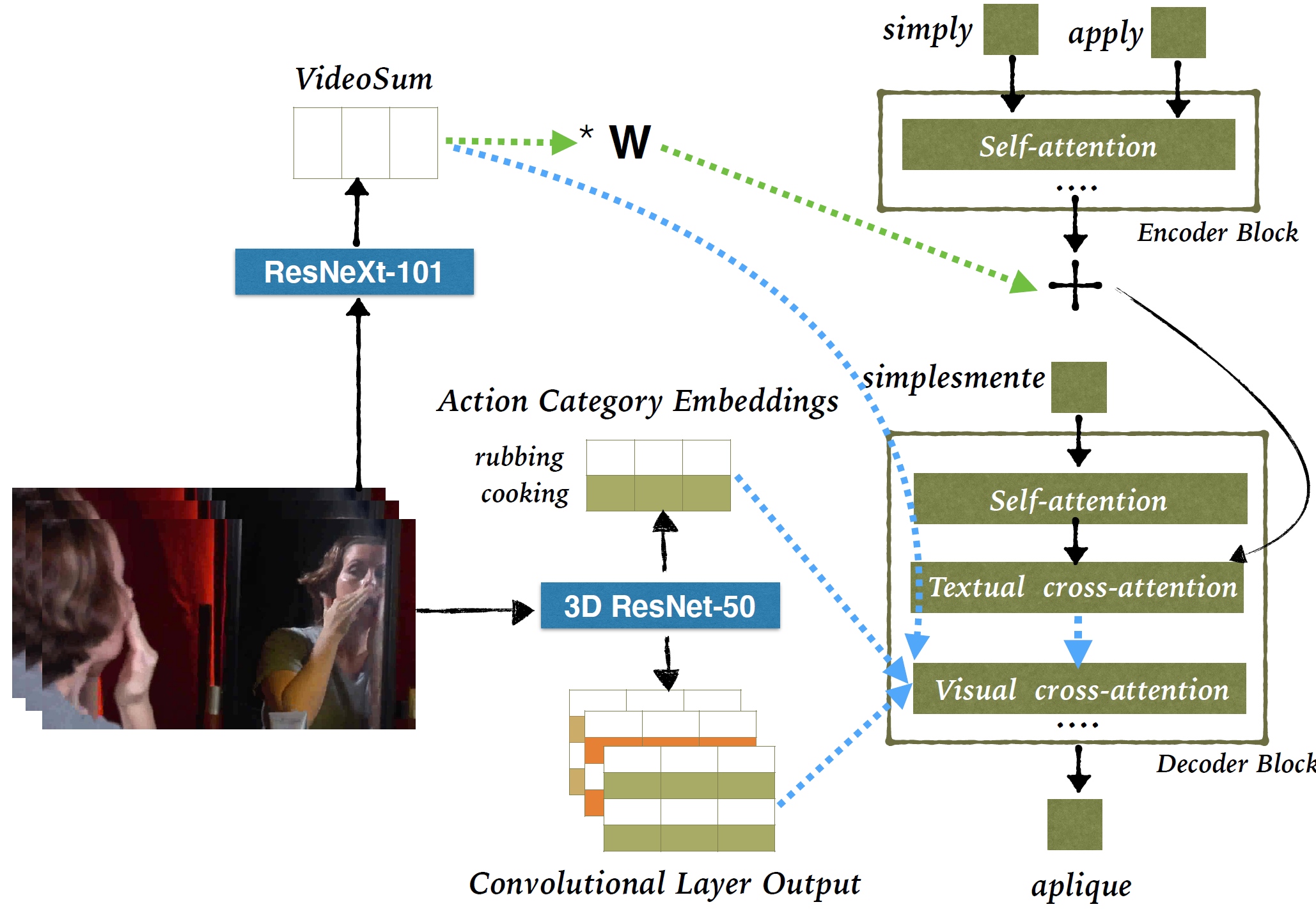}
\caption{Unimodal and multimodal transformers: \trans-\avc and the \trans-\avf extend the text-only \trans-\textonly with dashed green- and blue-arrow routes, respectively. Each multimodal model activates either the dashed green-arrow route for \trans-\avc or one of the three dashed blue-arrow routes (i.e. VideoSum, Action Category Embeddings or Convolutional Layer Output, as shown) for \trans-\avf.}
\label{fig:transformers}
\end{figure*}

The vanilla text-only transformer (\trans-\textonly) is used as a baseline, and we design two variants: with additive visual conditioning (\trans-\avc) and with attention to visual features (\trans-\avf). A \trans-\avc features a \enc-\avc and a vanilla transformer decoder (\dec-\vanilla), therefore utilising visual information only at the encoder side.  In contrast, a \trans-\avf is configured with a \enc-\vanilla and a \trans-\dec-\avf, exploiting visual cues only at the decoder. Figure~\ref{fig:transformers} summarises the two approaches.

\subsubsection{Multimodal Deliberation}
\label{multimodal deliberation}

\begin{figure*}[ht]
\centering
\includegraphics[width=.8\textwidth]{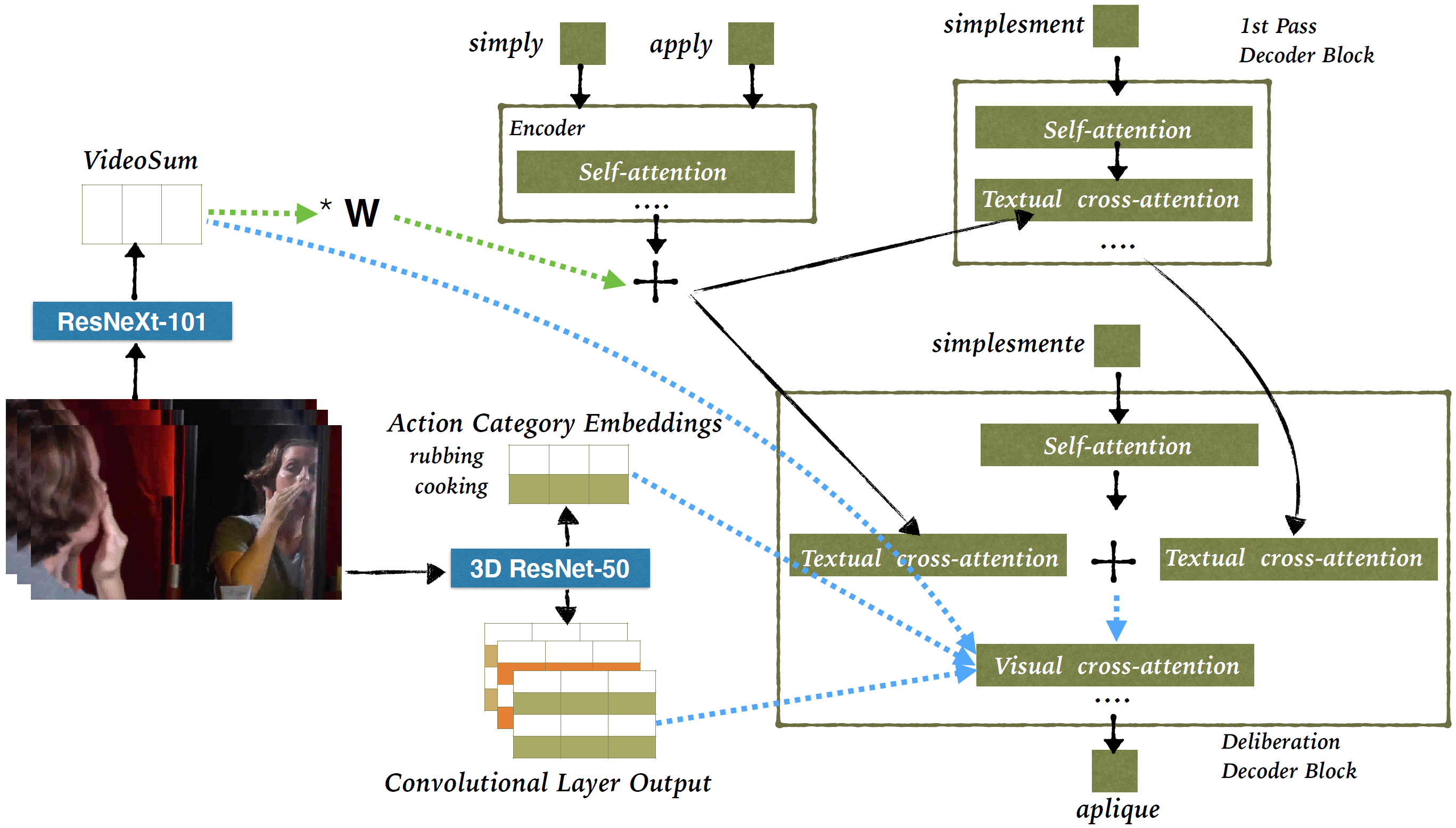}
\caption{Unimodal and multimodal \textbf{additive} deliberation: 
\delib-\avc and \delib-\avf extend the text-only \delib-\textonly with dashed green- and blue-arrow routes, respectively. Each multimodal model activates either the dashed green-arrow route for \delib-\avc or one of the three dashed blue-arrow routes (i.e. VideoSum, Action Category Embeddings or Convolutional Layer Output, as shown) for \delib-\avf.}
\label{fig:additive deliberation}
\end{figure*}

\begin{figure*}[ht!]
\centering
\includegraphics[width=.8\textwidth]{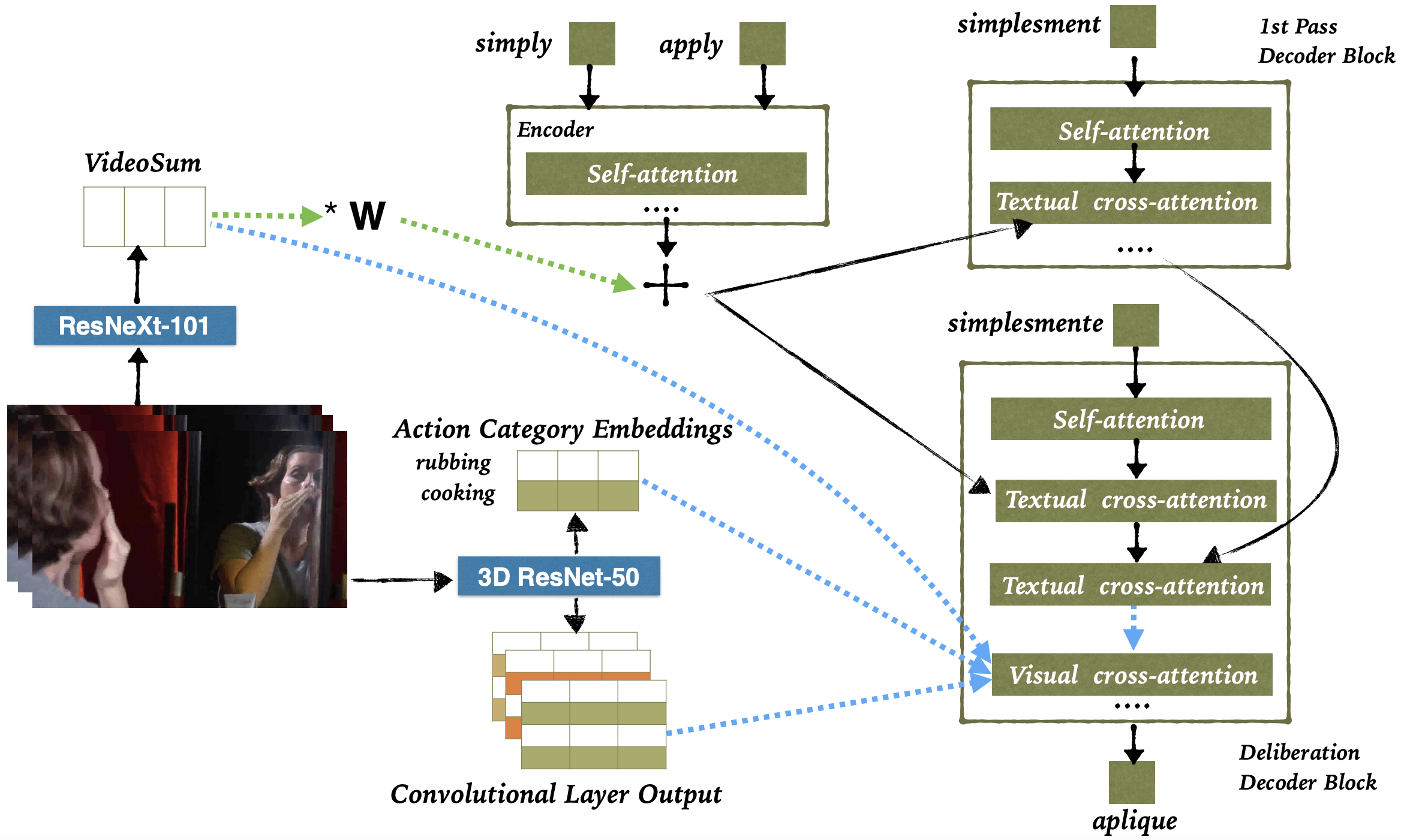}
\caption{Unimodal and multimodal \textbf{cascade} deliberation:
\delib-\avc and \delib-\avf extend the text-only \delib-\textonly with dashed green- and blue-arrow routes, respectively. Each multimodal model activates either the dashed green-arrow route for \delib-\avc or one of the three dashed blue-arrow routes (i.e. VideoSum, Action Category Embeddings or Convolutional Layer Output, as shown) for \delib-\avf.}
\label{fig:cascade deliberation}
\end{figure*}

Our multimodal deliberation models differ from each other in two ways: whether to use additive (\additive) \cite{ive2019distil} or cascade (\cascade) \textbf{textual} deliberation to integrate the \textbf{textual} attention to the original input and to the first pass, and whether to employ \textbf{visual} attention (\delib-\avf) or additive \textbf{visual} conditioning (\delib-\avc) to integrate the \textbf{visual} features into the textual MT model. Figures~\ref{fig:additive deliberation} and \ref{fig:cascade deliberation} show the configurations of our additive and cascade deliberation models, respectively, each also showing the connections necessary for \delib-\avc and \delib-\avf.\par

\smallsection{Additive (\additive) \& Cascade (\cascade) Textual Deliberation}

In an \textbf{additive-deliberation} second-pass decoder (\additive-\delib-\secondpass) block, the first layer is still self-attention, whereas the second layer is the addition of two separate attention sub-layers. The first sub-layer attends to the encoder output in the same way \dec-\vanilla does, while the attention of the second sub-layer is distributed across the concatenated first pass outputs and hidden states. The input to both sub-layers is the output of the self-attention layer, and the outputs of the sub-layers are summed as the final output and then (with a residual connection) fed to the visual attention layer if the decoder is multimodal or to the fully connected layer otherwise.

For the \textbf{cascade} version, the only difference is that, instead of two sub-layers, we have two separate, \textbf{successive} layers with the same functionalities.

It is worth mentioning that we introduce the attention to the first pass only at the initial three decoder blocks out of the total six of the second pass decoder (\delib-\secondpass), following \cite{ive2019distil}.\par

\smallsection{Additive Visual Conditioning (\delib-\avc) \& Visual Attention (\delib-\avf)}

\delib-\avc and \delib-\avf are simply applying \enc-\avc and \dec-\avf respectively to a deliberation model, therefore more details have been introduced in Section~\ref{multimodal encoding and decoding}.

For \delib-\avc, similar to in \trans-\avc, we add a projection of the visual features to the output of \enc-\vanilla, and use \dec-\vanilla as the first pass decoder and either additive or cascade deliberation as the \delib-\secondpass.\par

For \delib-\avf, in a similar vein as \trans-\avf, the encoder in this setting is simply \enc-\vanilla and the first pass decoder is just \dec-\vanilla, but this time \delib-\secondpass is responsible for attending to the first pass output as well as the visual features. For both additive and cascade deliberation, a visual attention layer is inserted immediately before the fully-connected layer, so that the penultimate layer of a decoder block now attends to visual information.


\section{Experiments}
\label{experiments}

\subsection{Dataset}
\label{dataset}
We stick to the default training/validation/test splits and the pre-extracted speech features for the How2 dataset, as provided by the organizers.
As for the \textbf{pre-processing}, we lowercase the sentences and then tokenise them using Moses \cite{moses}. We then apply subword segmentation \cite{sennrich2015neural} by learning separate English and Portuguese models with 20,000 merge operations each. The English corpus used when training the subword model consists of both the ground-truth video subtitles and the noisy transcripts produced by the underlying ASR system. We do not share vocabularies between the source and target domains. Finally for the \textbf{post-processing} step, we merge the subword tokens, apply recasing and detokenisation. The recasing model is a standard Moses baseline trained again on the parallel How2 corpus.

The baseline ASR system is trained on the How2 dataset as well. This system is then used to obtain noisy transcripts for the whole dataset, using beam-search with beam size of 10. The pre-processing pipeline for the ASR is different from the MT pipeline in the sense that the punctuations are removed and the subword segmentation is performed using SentencePiece \cite{sentencepiece} with a vocabulary size of 5,000. The test-set performance of this ASR is around 19\% WER.



\subsection{Training}
\label{training}
We train our transformer and deliberation models until convergence largely with \texttt{transformer\_big} hyperparameters: 16 attention heads, 1024-D hidden states and a dropout of 0.1. During inference, we apply beam-search with beam size of 10. For deliberation, we first train the underlying transformer model until convergence, and use its weights to initialise the encoder and the first pass decoder. After freezing those weights, we train \delib-\secondpass until convergence. The reason for the partial freezing is that our preliminary experiments showed that it enabled better performance compared to updating the whole model.
Following \cite{xia2017deliberation}, we obtain 10-best samples from the first pass with beam-search for source augmentation during the training of \delib-\secondpass.
 
We train all the models on an Nvidia RTX 2080Ti with a batch size of 1024, a base learning rate of 0.02 with 8,000 warm-up steps for the Adam \cite{kingma2014adam} optimiser, and a patience of 10 epochs for early stopping based on \texttt{approx-BLEU} (\tensortwotensor) for the transformers and 3 epochs for the deliberation models. After the training finishes, we evaluate all the checkpoints on the validation set and compute the real \bleu \cite{papineni2002bleu} scores, based on which we select the best model for inference on the test set.
The transformer and the deliberation models are based upon the \tensortwotensor\footnote{\url{https://github.com/tensorflow/tensor2tensor}} library \cite{vaswani2018tensor2tensor} (v1.3.0 RC1) as well as the vanilla transformer-based deliberation\footnote{\url{https://github.com/ustctf/delibnet}}~\cite{xia2017deliberation} and their multimodal variants\footnote{\url{https://github.com/ImperialNLP/MMT-Delib}}~\cite{ive2019distil}.

\section{Results \& Analysis}
\label{results and analysis}

\subsection{Quantitative Results}
\label{scores}
We report tokenised results obtained using the \textit{multeval} toolkit \cite{clark-etal-2011-better}. We focus on single system performance and thus, do not perform any ensembling or checkpoint averaging.

The \bleu scores of the models are shown in Table~\ref{results table}. Evident from the table is that the best models overall are \trans-\textonly and \trans-\avf-\clo with a \bleu score of 39.8, and the other multimodal transformers have slightly worse performance, showing score drops around 0.1. Also, none of the multimodal transformer systems are significantly different from the baseline, which is a sign of the limited extent to which visual features affect the output.

For additive deliberation (\additive-\delib), the performance variation is considerably larger: \avf-\videosum and \textonly take the lead with 37.6 \bleu, but the next best system (\avf-\clo) plunges to 37.2. The other two (\avc-\videosum \& \avf-\thace) also have noticeably worse results (36.0 and 37.0). Overall, however, \additive-\delib is still similar to the transformers in that the baseline generally yields higher-quality translations.

Cascade deliberation, on the other hand, is different in that its text-only baseline is outperformed by most of its multimodal counterparts. Multimodality enables boosts as large as around 1 \bleu point in the cases of \avf-\videosum and \avf-\thace, both of which achieve about 37.4 \bleu and are significantly different from the baseline.

\begin{table}[t!]
\caption{\bleu scores for the test set: bold highlights our best results. $\dagger$ indicates a system is significantly different from its text-only counterpart (p-value $\leq$ 0.05).}
\label{results table}
\vskip 0.15in
\begin{center}
\begin{sc}
\begin{tabular}{llll}
\toprule
Setup            & \trans    & \additive-\delib     & \cascade-\delib \\
\midrule
\textonly        & \bf 39.8  & \bf 37.6             & 36.4                  \\
\midrule
\avc-\videosum   & 39.7      & 36.0 $\dagger$       & 36.2                  \\
\avf-\videosum   & 39.7      & \bf 37.6             & \bf 37.4 $\dagger$    \\
\avf-\thace      & 39.7      & 37.0 $\dagger$       & 37.3 $\dagger$    \\
\avf-\clo        & \bf 39.8  & 37.2                 & 37.0                  \\
\bottomrule
\end{tabular}
\end{sc}
\end{center}
\end{table}

Another observation is that the deliberation models as a whole lead to worse performance than the canonical transformers, with \bleu deterioration ranging from 2.3 (across \avf-\videosum variants) to 3.5 (across \avc-\videosum systems), which defies the findings of \cite{ive2019distil}. We leave this to future investigations.

\subsection{Incongruence Analysis}
\label{incongruence analysis}

To further probe the effect of multimodality, we follow the incongruent decoding approach~\cite{elliott2018adversarial}, where our multimodal models are fed with mismatched visual features. The general assumption is that a model will have learned to exploit visual information to help with its translation, if it shows substantial performance degradation when given wrong visual features. The results are reported in Table~\ref{incongruency table}.

\begin{table}[t]
\caption{Incongruent decoding results for the test set: \bleu changes are w.r.t the congruent counterparts from Table~\ref{results table}. $\dagger$ marks incongruent decoding results that are significantly different (p-value $\leq$ 0.05) from congruent counterparts.}
\label{incongruency table}
\vskip 0.15in
\begin{center}
\begin{sc}
\begin{tabular}{llll}
\toprule
Setup            & \trans    & \additive-\delib  & \cascade-\delib \\
\midrule
\avc-\videosum   & $\downarrow$ 0.5 $\dagger$     & $\uparrow$ 0.1      & $\downarrow$ 0.6 $\dagger$     \\
\avf-\videosum   & $\downarrow$ 0.3      & 0      & $\downarrow$ 0.1     \\
\avf-\thace      & $\downarrow$ 0.4 $\dagger$     & $\uparrow$ 0.1   & $\downarrow$ 0.2 $\dagger$     \\
\avf-\clo     & $\downarrow$ 0.1      & $\downarrow$ 0.2   & $\downarrow$ 0.2    \\
\bottomrule
\end{tabular}
\end{sc}
\end{center}
\end{table}

Overall, there are considerable parallels between the transformers and the cascade deliberation models in terms of the incongruence effect, such as universal performance deterioration (ranging from 0.1 to 0.6 \bleu) and more noticeable score changes ($\downarrow$ 0.5 \bleu for \trans-\avc-\videosum and $\downarrow$ 0.6 \bleu for \cascade-\delib-\avc-\videosum) in the \avc-\videosum setting compared to the other scenarios. Additive deliberation, however, manifests a drastically different pattern, showing almost no incongruence effect for \avf-\videosum, only a 0.2 \bleu decrease for \avf-\clo, and even a 0.1 \bleu boost for \avf-\thace and \avc-\videosum.\par

Therefore, the determination can be made that \trans and \cascade-\delib models are considerably more sensitive to incorrect visual information than \additive-\delib, which means the former better utilise visual clues during translation.\par

Interestingly, the extent of performance degradation caused by incongruence is not necessarily correlated with the congruent \bleu scores. For example, \trans-\avf-\clo is on par with \trans-\avf-\thace in congruent decoding (differing by around 0.1 \bleu), but the former suffers only a 0.1-\bleu loss with incongruence whereas the figure for the latter is 0.4, in addition to the fact that the latter becomes significantly different after incongruent decoding. This means that some multimodal models that are sensitive to incongruence likely complement visual attention with textual attention but without getting higher-quality translation as a result.

The differences between the multimodal behaviour of additive and cascade deliberation also warrant more investigation, since the two types of deliberation are identical in their utilisation of visual features and only vary in their handling of the textual attention to the outputs of the encoder and the first pass decoder.\par

\section{Conclusions}
\label{conclusions}
We explored a series of transformers and deliberation based models to approach cascaded multimodal speech translation as our participation in the How2-based speech translation task of IWSLT 2019. We submitted the \trans-\avf-\clo system, which is a canonical transformer with visual attention over the convolutional features, as our primary system with the remaining ones marked as contrastive ones. The primary system obtained a \bleu of 39.63 on the public IWSLT19 test set, whereas \trans-\textonly, the top contrastive system on the same set, achieved 39.85. Our main conclusions are as follows: (i) the visual modality causes varying levels of translation quality damage to the transformers and additive deliberation, but boosts cascade deliberation; (ii) the multimodal transformers and cascade deliberation show performance degradation due to incongruence, but additive deliberation is not as affected; (iii) there is no strict correlation between incongruence sensitivity and translation performance.

\section{Acknowledgements}
\label{acknowledgement}
This work was supported by the MultiMT (H2020 ERC Starting Grant No. 678017) and MMVC (Newton Fund Institutional Links Grant, ID 352343575) projects.

\bibliographystyle{IEEEtran}
\bibliography{main}
\end{document}